\documentclass[conference]{IEEEtran}
\IEEEoverridecommandlockouts
\usepackage{cite}
\usepackage{amsmath,amssymb,amsfonts}
\usepackage{algorithmic}
\usepackage{graphicx}
\usepackage{subfigure}
\usepackage{textcomp}
\usepackage{xcolor}
\usepackage{float}
\def\BibTeX{{\rm B\kern-.05em{\sc i\kern-.025em b}\kern-.08em
    T\kern-.1667em\lower.7ex\hbox{E}\kern-.125emX}}
\usepackage{booktabs}
\usepackage[square,comma,numbers]{natbib}
\begin{document}

\title{Synaptic Scaling and Optimal Bias Adjustments for Power Reduction in Neuromorphic Systems}
% {\footnotesize \textsuperscript{*}Note: Sub-titles are not captured in Xplore and
% should not be used}
% \thanks{Identify applicable funding agency here. If none, delete this.}

\author{\IEEEauthorblockN{Cory Merkel}
\IEEEauthorblockA{Brain Lab, 
Rochester Institute of Technology\\
Rochester, NY, USA \\
cemeec@rit.edu}
}

\maketitle

\begin{abstract}
Recent animal studies have shown that biological brains can enter a low power mode in times of food scarcity.  This paper explores the possibility of applying similar mechanisms to a broad class of neuromorphic systems where power consumption is strongly dependent on the magnitude of synaptic weights.  In particular, we show through mathematical models and simulations that careful scaling of synaptic weights can significantly reduce power consumption (by over 80\% in some of the cases tested) while having a relatively small impact on accuracy.  These results uncover an exciting opportunity to design neuromorphic systems for edge AI applications, where power consumption can be dynamically adjusted based on energy availability and performance requirements.
\end{abstract}


\section{Introduction}

Neuromorphic computing is gaining traction as a promising paradigm for achieving artificial intelligence (AI) on energy-constrained edge devices \cite{hendy2022review,roy2019towards}.  One of the key design principles of neuromorphic systems involves the abstraction and application of biological mechanisms for efficient computation and adaptation \cite{Mead1989}.  Therefore, much of the work in the area of neuromorphic computing has focused on efficient bio-inspired designs of neurons, synapses, communication strategies, and learning rules.  However, in edge devices, it is also critical to have efficient and dynamic management of resources, such as energy.  Some methods have been proposed in this area, such as early exit neural networks \cite{bolukbasi2017adaptive,merkel2020exploring} and dynamic voltage frequency scaling \cite{hoppner2019dynamic}, but these techniques lack direct biological motivation.

In this paper, we propose a new bio-inspired method that could be used to dynamically scale the power consumption of neuromorphic systems.  First, we note that a significant portion ($>50\%$) of the energy consumed in the brain's gray matter is tied to synaptic activity \cite{harris2012synaptic}.  Interestingly, a broad class of neuromorphic systems, such as those based on non-volatile memory (NVM) crossbars \cite{burr2017neuromorphic}, also exhibit a strong dependence between power consumption and synaptic activity.  In a recent animal study \cite{padamsey2022neocortex}, it was shown that biological systems seem to have evolved a clever method for leveraging this strong correlation between the brain's energy consumption and the strength of synaptic connections between neurons.  In times of food scarcity, it was found that the brain reduces adenosine triphosphate (ATP) usage by reducing neurotransmitter receptor conductance at the post-synaptic neuron.  In order for the neuron to maintain the same level of activity, it also depolarizes its resting potential, as well as reducing its membrane resistance.  A key insight in this paper is that these two complementary actions are analogous to scaling the magnitude of weights and adjusting the bias of neurons, respectively.  We show that applying these two actions to the weight and biases of neuromorphic systems like those based on NVMs could be a promising solution for dynamic scaling of energy usage.

The rest of this paper proceeds as follows:  Section \ref{sec:theory} provides a detailed theoretical motivation for this work.  Section \ref{sec:single} gives results of the proposed energy scaling method for single neurons and single-layer neural networks.  Section \ref{sec:multi} discusses results related to multilayer neural networks.  Section \ref{sec:conc} concludes this work.

\section{Theoretical Motivation}
\label{sec:theory}

\begin{figure}[!t]
    \centering
    \includegraphics[width=0.6\columnwidth]{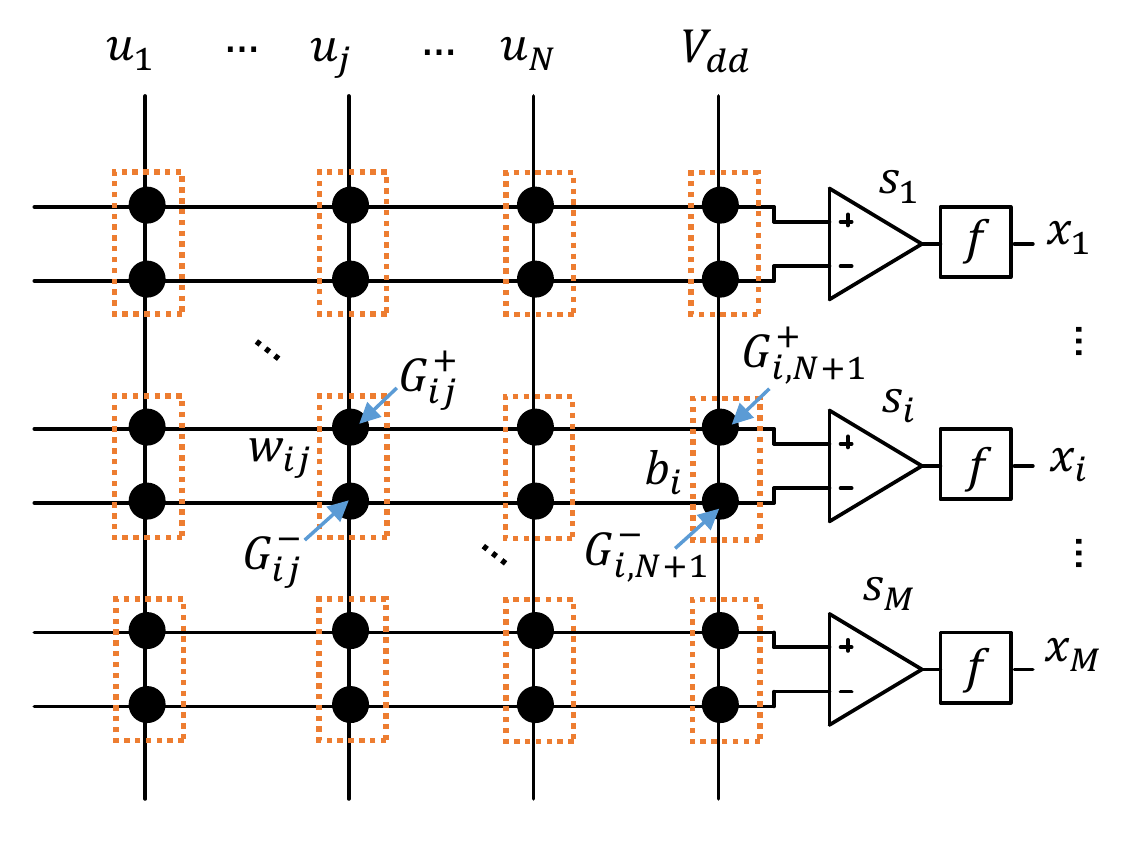}
    \caption{Neuromorphic NVM crossbar for implementing a layer of neurons.}
    \label{fig:xbar}
    \vspace{-4mm}
\end{figure}

A neuromorphic implementation of a layer of neurons using an NVM crossbar is shown in Figure \ref{fig:xbar} \cite{zhang2018neuromorphic}.  The weight matrix is formed by the conductances of several NVM devices.  There are many choices for the NVM technology that is used, such as resistive random access memory (ReRAM), phase change memory (PCRAM), ferroelectric devices, etc. \cite{upadhyay2019emerging}.  The crossbar exploits Ohm's Law and Kirchoff's Current Law to perform matrix-vector multiplication.  When the inputs $u_{j}$ and outputs $x_{i}$ are represented as voltages normalized to $V_{dd}$, then 
\begin{equation}
x_{i}=f\left(\sum\limits_{j}u_{j}(G_{ij}^{+}-G_{ij}^{-})+(G_{i,N+1}^{+}-G_{i,N+1}^{-})\right)
\label{eqn:xbar}
\end{equation}
where $f$ is the activation function.  The weights and biases are formed by the differences of two conductances (also considered to be normalized to some maximum conductance value in this case) in order to achieve both positive and negative values.  The subtraction is represented generically here with a (current-mode) unity gain amplifier, but equivalent behavior could be achieved using other implementations.

Simplifying the notation, we can now write the output of a particular neuron $x$ in the familiar way:
\begin{equation}
x=f(s)=f(\mathbf{u}\cdot\mathbf{w}+b)
\end{equation}
where $\mathbf{w}$ and $b$ are formed by the conductance pairs in the two rows of the crossbar associated with the neuron.  Note that $\mathbf{w}$ and $\mathbf{u}$ are vectors of length $N$.  Here, we will make a simplifying assumption that positive weight or bias values will have $G_{ij}^{-}\approx 0$ and negative weight or bias values will have $G_{ij}^{+}\approx 0$.  This means that the magnitude of a weight will be approximately equal to the sum of the associated conductances.  With this assumption, we can easily write the synaptic power consumption associated with a neuron as
\begin{equation}
SP\propto(\mathbf{u}\circ\mathbf{u})\cdot|\mathbf{w}|+|b|
\end{equation}
where $\circ$ is the element-wise product (Hadamard product), and $|\cdot|$ is the element-wise absolute value.

%Now, the total steady state current flowing through the crossbar can be written as
%\begin{equation}
%i_{total} = \sum\limits_{j=1}^{N}v_{uj}\sum\limits_{i=1}^{M}G_{ij}^{+}+G_{ij}^{-}= %\sum\limits_{j=1}^{N}v_{uj}^{l-1}G_{j},
%\end{equation}
%where we define $G_{j}$ as the sum of conductances connected to the $j^{\mathrm{th}}$ input.  The unknown values of $G_{j}$ can be determined through several observations of $i_{total}$ for different input voltages.  For example, setting $v_{u1}=V_{dd}$ and grounding all other inputs leads to $G_{1}=i_{total}/V_{dd}$.  The weight value $w_{ij}$ associated with a given pair of neurons $i$ and $j$ is proportional to $G_{ij}^{+}-G_{ij}^{-}$.  This means that there are several ways (potentially infinite) to achieve a given weight with different values of the two conductances.  However, we assume here that for positive weights, $G_{ij}^{-}\approx 0$ and for negative weights, $G_{ij}^{+}\approx 0$.  This is a safe assumption since it will lead to the lowest power consumption for a given weight matrix implementation, which is a primary goal, especially in edge devices.  It also means that each weight will have a one-one mapping to conductance values.  Now, we can write
%\begin{equation}
%|w_{ij}| \propto G_{ij}^{+}+G_{ij}^{-}
%\end{equation}
%If $\mathbf{u}$ is known, then by measuring the crossbar current (also referred to in this paper as the power information), we can determine the 1-norm of each column of $\mathbf{W}$.  The utility of this information depends on whether we have access to the crossbar outputs and activation function $f$.

\begin{table}[!t]
\centering
\caption{Different activation functions and the associated values of optimal bias adjustment for $p(s)=\frac{1}{\sigma}\phi\left(\frac{s-\mu}{\sigma}\right)$ and $\frac{\partial\mathcal{L}}{\partial x}=1$.}
\label{tab:minb}
\begin{tabular}{lll} 
\toprule
Activation\\ Function & $\Delta b^{*}$ & Parameters  \\ 
\hline
Linear & $(1-\epsilon)(\mu-b)$ &   -          \\
ReLU &  $(1-\epsilon)\left[\mu+\sigma\frac{\phi\left(\frac{-\mu}{\sigma}\right)}{1-\Phi\left(\frac{-\mu}{\sigma}\right)}-b\right]$ & - \\
Sigmoid & $(1-\epsilon)(B\mu-b)$                                     &     $B=\frac{1.05^{2}}{(\sigma^{2}+1.05^{2})}$        \\
tanh & $(1-\epsilon)(B\mu-b)$                                      &   $B=\frac{0.75^{2}}{(\sigma^{2}+0.75^{2})}$          \\
Step                   & $(\epsilon-1)b$                                & -             \\
\bottomrule
\end{tabular}
\end{table}

For large fan-in neurons, this will be dominated by the first term, so the power could easily be scaled by reducing the magnitudes of the weights by a factor $\epsilon$:
\begin{equation}
\tilde{x}=f(\tilde{s})=f(\epsilon\mathbf{u}\cdot\mathbf{w}+b+\Delta b)
\end{equation}
However this will change the neuron output, leading to increased loss.  One way to combat this is to adjust the bias by $\Delta b$ to compensate for the effect of the weight scaling.  Now the goal is, given an $\epsilon$, find $\Delta b$ to minimize the effect on the loss.  For simplicity, we assume that we are at a local minimum in the weight space so the loss will be monotonically increasing if we move only a small amount from the original weight values.  The optimal value of the bias adjustment can be estimated for small values of $\epsilon$ as (see Appendix):
\begin{equation}
\Delta b^{*}\approx (1-\epsilon)\left[\frac{\mathbb{E}\left[\delta^{2} s\right]}{\mathbb{E}\left[\delta^{2}\right]}-b\right]
\label{eqn:minb}
\end{equation}
Here, $\mathbb{E}[\cdot]$ is the expected value and $\delta\equiv\partial \mathcal{L}/\partial s$ has the usual definition from the backpropagation algorithm, where $\mathcal{L}$ is the loss.  Table \ref{tab:minb} shows several values of $\Delta b^{*}$ for different activation functions when $p(s)=\frac{1}{\sigma}\phi(\frac{s-\mu}{\sigma})$, where $p(s)$ is the distribution of neuron inputs, $\phi$ is the standard normal distribution function, $\mu$ is the mean, and $\sigma$ is the standard deviation.  The standard cumulative distribution function is written as $\Phi$. 
 Also, note that the results in Table \ref{tab:minb} assume a linear relationship between the loss and the neuron's output.  More discussion on how these are derived is found in the Appendix.

\begin{figure}
\subfigure[]{
\includegraphics[width=0.485\columnwidth]{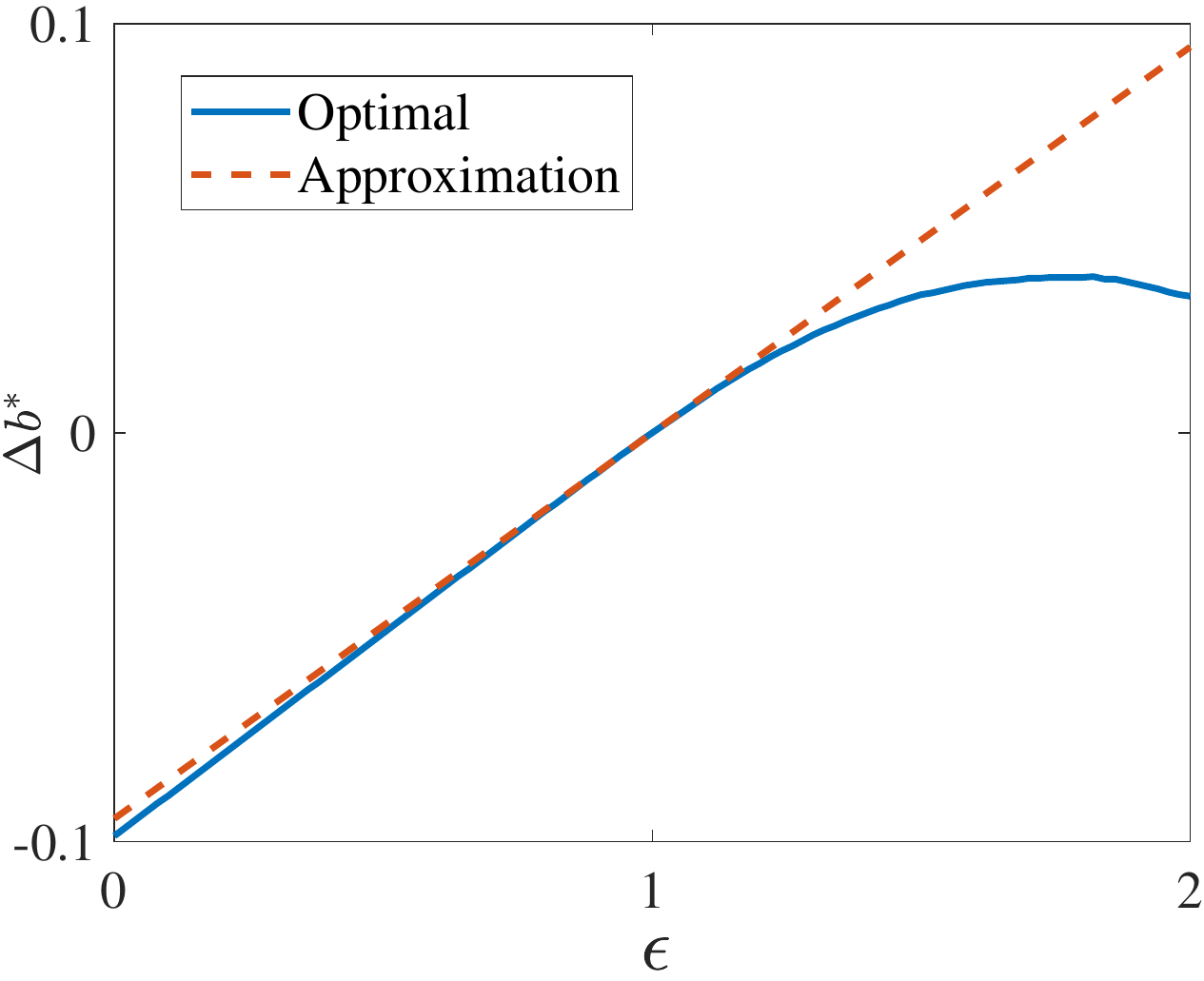}
}
\hspace{-5mm}
\subfigure[]{
\includegraphics[width=0.485\columnwidth]{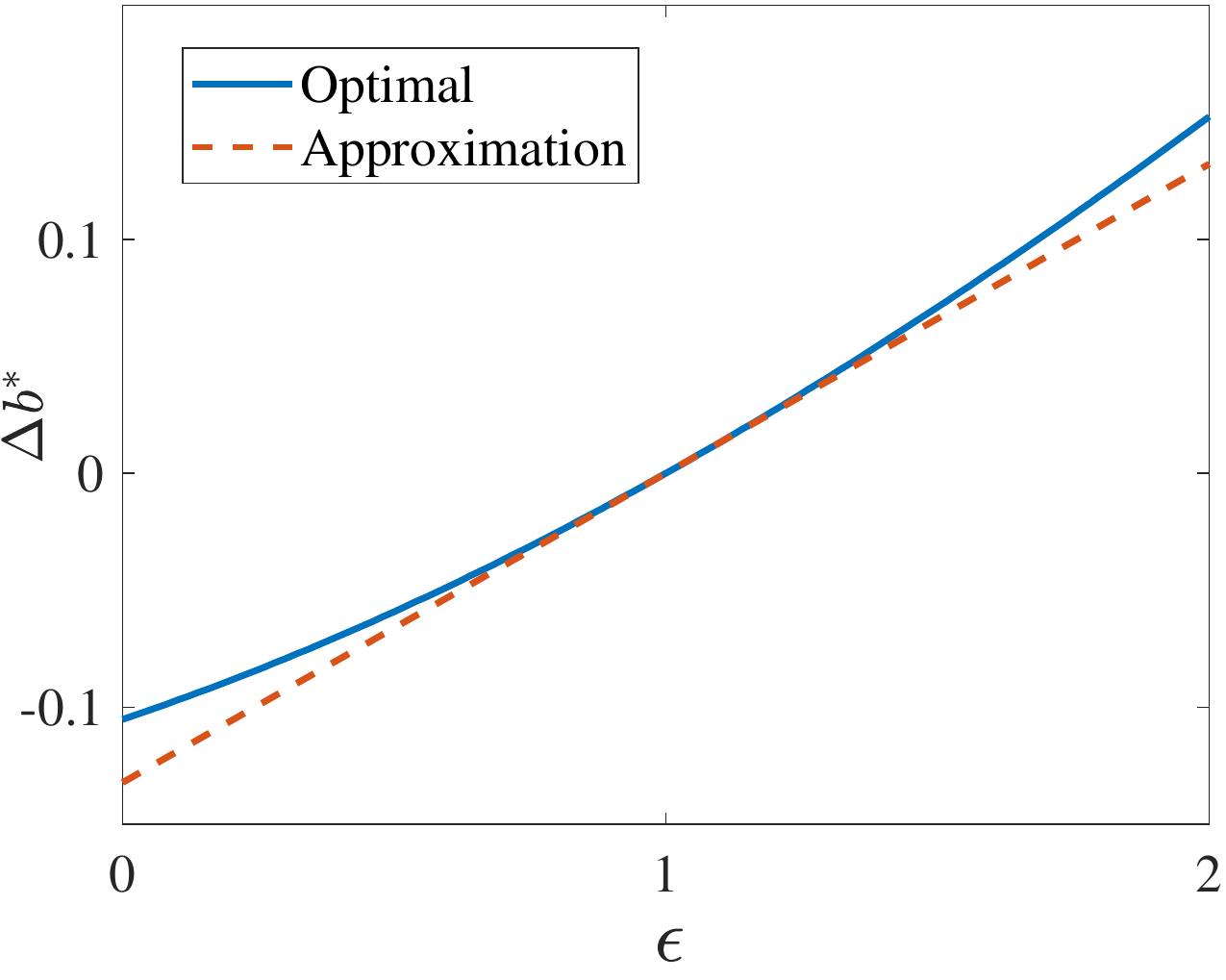}
}
\caption{Examples of optimal bias adjustment for a single neuron with (a) ReLU and (b) sigmoid activation function.  For both cases, $\mu=-0.1$, $\sigma=0.25$, and $b=0.7$.}
\label{fig:optbias}
 \end{figure}

Examples of the true optimal bias adjustment vs. our  model approximation for two different activation functions are shown in Figure \ref{fig:optbias}.  Here, we show $\epsilon$ ranging from 0 to 2, and we expect the approximation to match the optimal bias adjustment at $\epsilon=1$.  Indeed, for both ReLU and sigmoid activation functions, the approximation is tangent to the optimal bias adjustment for the unscaled weights ($\epsilon=1$).  Since our model is based on a first-order expansion of the loss, it is expected to provide an accurate approximation only for small deviations from $\epsilon=1$ and $\Delta b=0$.  In the case of the examples in Figure \ref{fig:optbias}, the approximation matches the true optimal value of $\Delta b$ fairly closely for the full range of interest ($\epsilon=0$ to $\epsilon=1$).  However, this will not always be the case.

\section{Single Neuron Analysis}
\label{sec:single}

\begin{figure}[!t]
\centering
\subfigure[]{
\includegraphics[width=0.55\columnwidth]{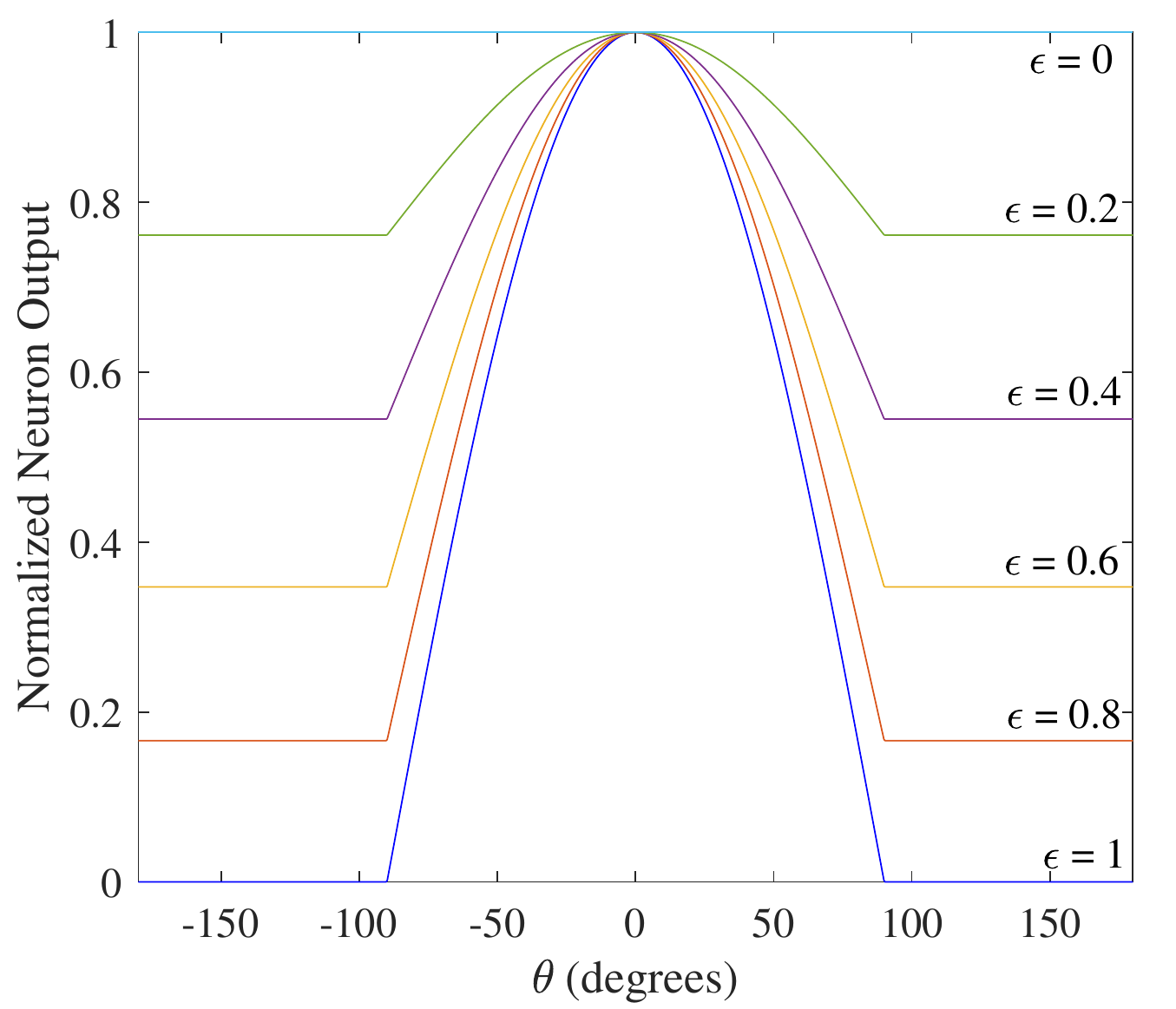}
}
\subfigure[]{
\includegraphics[width=0.55\columnwidth]{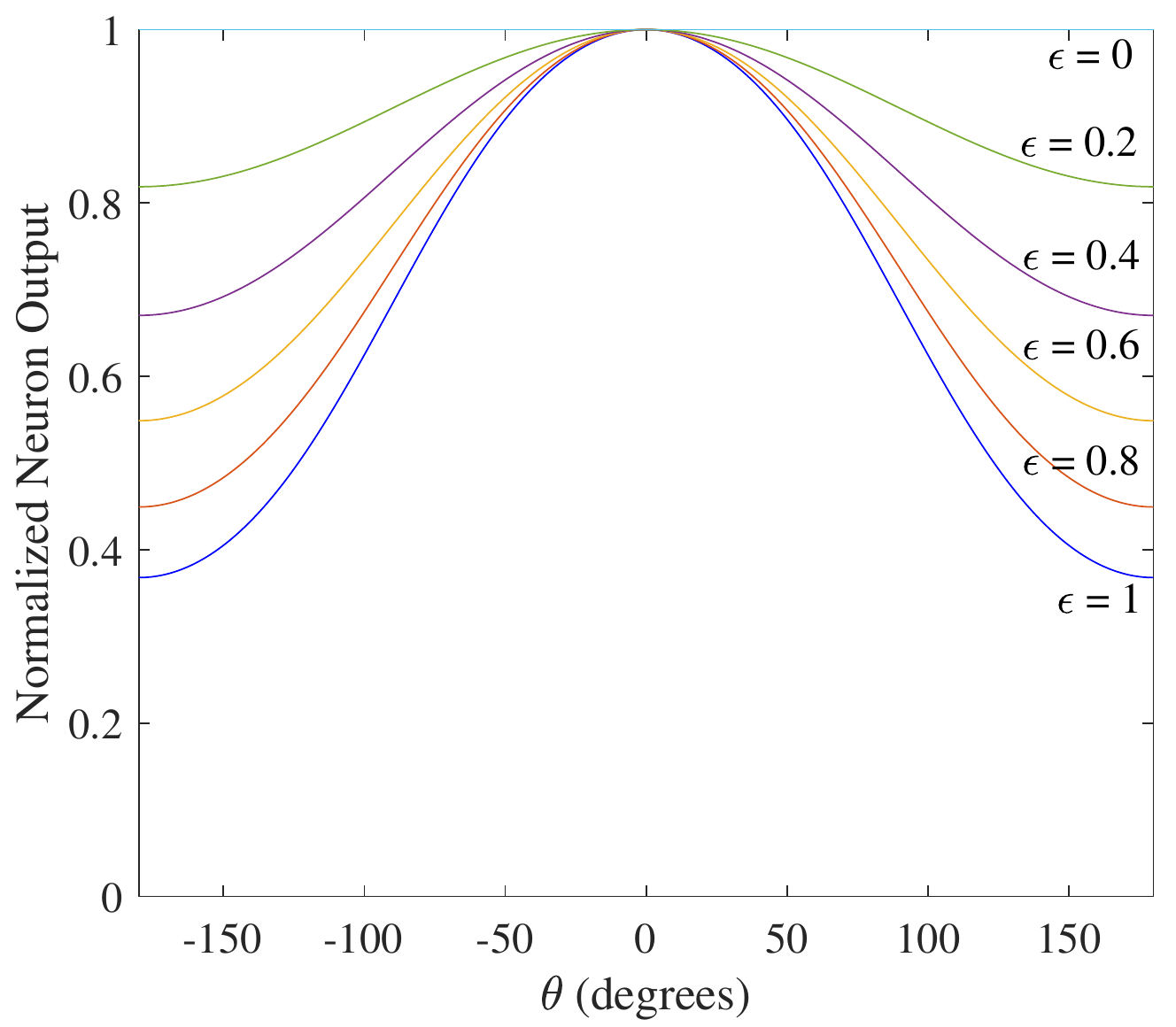}
}
\caption{Normalized neuron output vs. angle between $\mathbf{u}$ and $\mathbf{w}$ for (a) ReLU and (b) sigmoid activation functions.}
\label{fig:tuningcurves}
\end{figure}

Now, let us analyze the effect of weight scaling and bias adjustment on a neuron's behavior.  The neuron input can be written as

\begin{equation}
s=\epsilon\lVert\mathbf{u}\rVert\lVert\mathbf{w}\rVert\mathrm{cos}\theta+b+\Delta b,
\end{equation}
where $\theta$ is the angle between $\mathbf{u}$ and $\mathbf{w}$.  Figure \ref{fig:tuningcurves} shows the normalized output of neurons with ReLU and sigmoid activation functions when the norms of $\mathbf{u}$ and $\mathbf{w}$ are set to 1 and $\theta$ is varied from -180 to 180 degrees.  In each case, the bias is set to 0 and the optimal bias adjustments are made using Table \ref{tab:minb}, with $\mu=0$ and $\sigma=1$.  Note that as the value of $\epsilon$ becomes smaller the curves become wider, which matches recent observations in animal studies \cite{padamsey2022neocortex}.  This tuning curve broadening reduces the precision with which particular features are detected.

\begin{figure}[!ht]
\centering
\includegraphics[width=0.75\columnwidth]{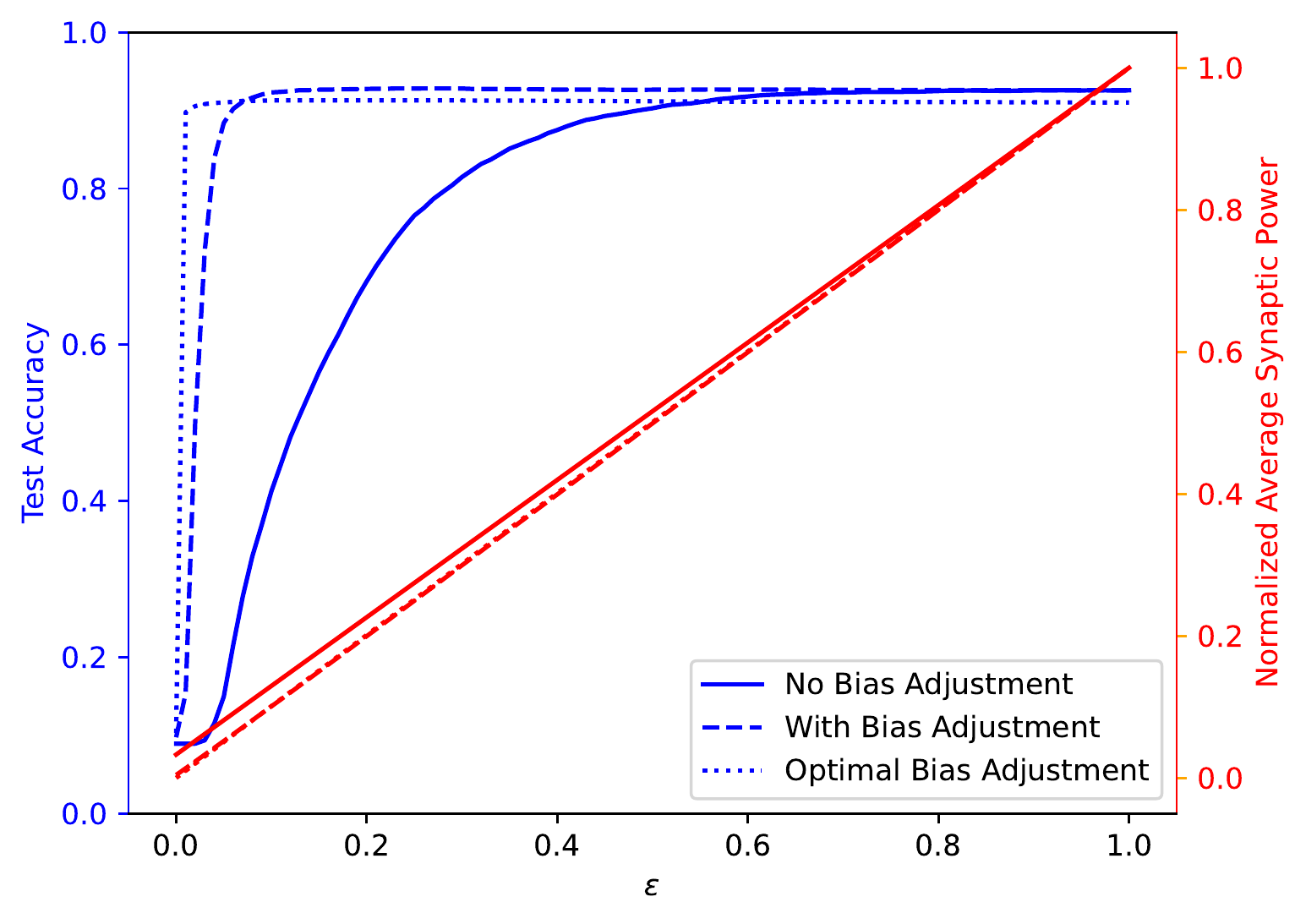}
\caption{Test accuracy vs. $\epsilon$ for a single-layer neural network with sigmoid activation function trained on the MNIST dataset.  The right $y$ axis shows the normalized average synaptic power.  For both axes, the solid line indicates results when no bias adjustment is applied, while the dashed line shows results for the bias adjustment.  The dotted line shows the bias adjustment found through backpropagation.}
\label{fig:mnistsinglelayer}
\end{figure}

The effect of broadening tuning curves on the performance of a single-layer neural network can be seen in Figure \ref{fig:mnistsinglelayer}.  Here, we trained a single-layer neural network with sigmoid activation function on the MNIST dataset \cite{deng2012mnist}.  First the network was trained for 25 epochs using the Adam optimizer, attaining a test accuracy of $\sim 92\%$.  Then, the model was evaluated for different values of $\epsilon$ with three different methods of bias adjustment applied to each of neurons.  In the first method, the bias is left constant at the trained value (solid blue curve).  It can be seen that the test accuracy starts to fall off when $\epsilon$ falls below about 0.5.  In the second method (blue dashed line), we apply the bias adjustment in Table \ref{tab:minb}.  In this case, the test accuracy remains approximately constant until $\epsilon$ drops below 0.1.  For the third method, we found the true optimal bias adjustment by tuning only the biases using backpropagation, while the weights were frozen at their trained values multiplied by $\epsilon$.  Note that the bias values are tuned only using the training dataset, while the test data is left out for evaluation.  This method yields the dotted blue curve, which shows a drop in accuracy only at $\epsilon$ values that are very close to 0.  Interestingly, the maximum accuracy of this method is slightly below (about 1\%) the other two methods, which is likely due to overfitting of the bias values to the training data.

In addition to test accuracy, we have also plotted the normalized average synaptic power (NASP), which we define as:
\begin{equation}
NASP\equiv\frac{\sum\limits_{i}\sum\limits_{p}\bigg[\epsilon(\mathbf{u}_{i}^{p}\circ\mathbf{u}_{i}^{p})\cdot|\mathbf{w}_{i}|+|b_{i}+\Delta b_{i}|\bigg]}{\sum\limits_{i}\sum\limits_{p}\bigg[(\mathbf{u}_{i}^{p}\circ\mathbf{u}_{i}^{p})\cdot|\mathbf{w}_{i}|+|b_{i}|\bigg]}
\end{equation}
where the summations are taken over all of the samples from the dataset ($p$ index) and all of the neurons in the network ($i$ index).  Without any bias adjustment, this is a linear function of $\epsilon$ (solid red curve).  However, we see that even with a bias adjustment, the relationship with $\epsilon$ is approximately linear (dashed and dotted red curves) due to the much larger influence of the weights compared to the influence of $\Delta b$.  For very small $\epsilon$ values the NASP of the networks with bias adjustment dips below the NASP of the unadjusted network.  This means that, on average the bias adjustment tends to decrease the overall magnitude of the bias.  However, the difference in NASP is fairly small, and the main conclusion is that weight scaling with bias adjustment is able to reduce NASP by about the same amount as the unadjusted case while maintaining much higher levels of accuracy at smaller values of $\epsilon$.

\begin{figure}[!ht]
\centering
\includegraphics[width=0.75\columnwidth]{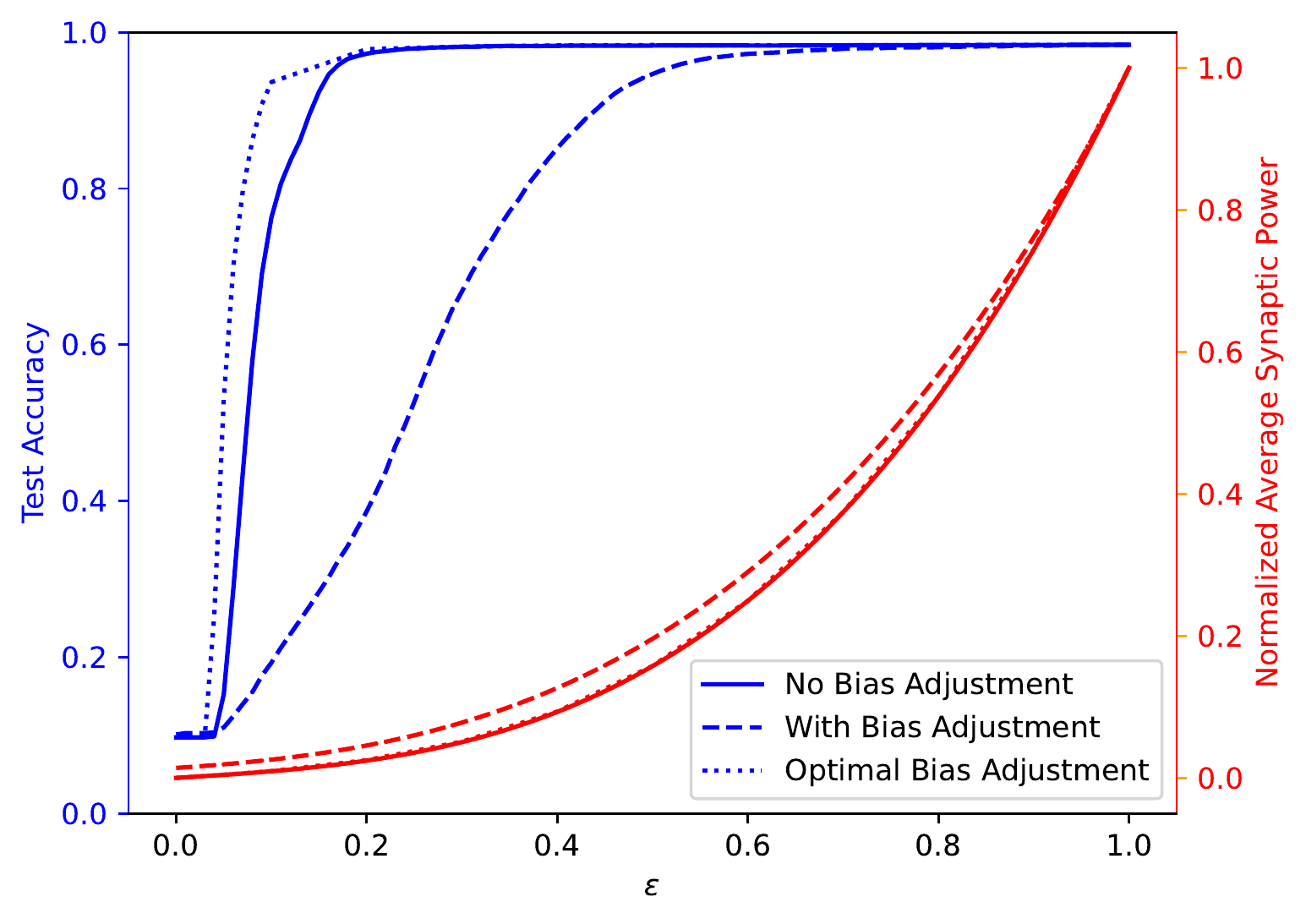}
\caption{Test accuracy vs. $\epsilon$ for a multilayer neural network with ReLU activation function in the hidden layer and sigmoid activation function in the output layer trained on the MNIST dataset.  The right $y$ axis shows the normalized average synaptic power.  For both axes, the solid line indicates results when no bias adjustment is applied, while the dashed line shows results for the bias adjustment.  The dotted line shows the bias adjustment found through backpropagation.}
\label{fig:mnistmlp}
\end{figure}

\section{Multilayer Neural Networks}
\label{sec:multi}

In this section, we discuss the application of the proposed synaptic scaling and bias adjustment on multilayer neural networks.  We trained a multilayer perception (MLP) with a single 1000-neuron ReLU hidden layer and 10 sigmoid output neurons on the MNIST dataset, using the same training parameters as in the previous section.  The results are shown in Figure \ref{fig:mnistmlp}.  First, we note that the maximum accuracy at $\epsilon=1$ is larger than that of the single-layer neural network.  This is expected, since the MLP is better able to learn complex features of the images.  The solid blue curve shows the effect of $\epsilon$ when no bias adjustment is added.  In this case the accuracy doesn't drop off until $\epsilon$ falls well below 0.2.  Moreover, we see that the test accuracy with bias adjustment (blue dashed curve) is worse than the accuracy without bias adjustment until $\epsilon$ becomes close to 1.  There are two likely causes of this.  First, any small model error in the hidden layer will be amplified by the output layer.  Second, the hidden layer bias adjustment will change the statistics of the inputs to the output layer.  This change wasn't captured in our simulations since we calculated the statistics based on the original network without any weight scaling ($\epsilon=0$).  However, if we scale the biases using backpropagation, as we did in the last section, then we are still able to get better test accuracy at lower $\epsilon$ values.  In addition, it is observed that the NASP is approximately equal among all 3 techniques.

\section{Conclusions}
\label{sec:conc}

This paper explored a bio-inspired method for scaling the power consumption of neuromorphic systems based on reduction of weight magnitudes and adjustment of neuron biases.  It was found that scaling weight magnitudes alone can have a large impact on power consumption with a small effect on the accuracy.  However, as weight magnitudes are scaled to extremely small values, bias adjustments are needed to maintain acceptable accuracy levels.  This paper only considered a rough model of the neuromorphic power consumption based on synaptic activity, and further work will be needed to verify the effectiveness of the technique on real neuromorphic hardware.

\section*{Appendix}
\subsection{Derivation of Optimal Bias Adjustment}
\begin{equation}
\Delta b^{*}=\underset{\Delta b}{\mathrm{arg\:min}}\:\: \mathcal{L}(\mathcal{D}_\mathrm{train};\epsilon\mathbf{w},b+\Delta b)
\end{equation}
Now expand the loss around $\epsilon=1$ and $\Delta b=0$ for a particular neuron:
\begin{equation}
\mathcal{L}=\mathcal{L}_{0}+\Delta\mathcal{L}\approx\mathcal{L}_{0}+(\epsilon-1)\frac{\partial\mathcal{L}}{\partial\epsilon}+\Delta b\frac{\partial\mathcal{L}}{\partial\Delta b}
\end{equation}
We want to minimize the expected change in the loss over the true data distribution:
\begin{equation}
\begin{aligned}
\mathbb{E}_{\mathcal{D}}[\Delta\mathcal{L}]&=\int\limits_{\mathcal{D}}p(s)\left[(\epsilon-1)\frac{\partial\mathcal{L}}{\partial x}\frac{\partial x}{\partial s}\frac{\partial s}{\partial \epsilon}+\Delta b\frac{\partial\mathcal{L}}{\partial x}\frac{\partial x}{\partial s}\frac{\partial s}{\partial \Delta b}\right]\mathrm{d}s&\\
&=\int\limits_{\mathcal{D}}p(s)\frac{\partial\mathcal{L}}{\partial x}f'(s)\left[(\epsilon-1)(s-b)+\Delta b\right]\mathrm{d}s &
\end{aligned}
\end{equation}
To minimize this expected value w.r.t. $\Delta b$, we set the $\frac{\partial \mathbb{E}[\Delta \mathcal{L}^{2}]}{\partial\Delta b}=0$ and solve for $\Delta b$:
\begin{equation}
\begin{aligned}
\Delta b^{*} = &(1-\epsilon)\left[\frac{\int p(s)\left[\frac{\partial\mathcal{L}}{\partial x}f'(s)\right]^{2}s\mathrm{d}s}{\int p(s)\left[\frac{\partial\mathcal{L}}{\partial x}f'(s)\right]^{2}\mathrm{d}s}-b\right] \\
= & (1-\epsilon)\left[\frac{\mathbb{E}\left[\delta^{2} s\right]}{\mathbb{E}\left[\delta^{2}\right]}-b\right]
\end{aligned}
\label{eqn:minb}
\end{equation}
\subsection{Specific Cases}
The following specific cases apply when $p(s)$ is a normal distribution function with mean $\mu$ and variance $\sigma^{2}$.
\subsubsection{Linear}
For a linear activation function, $f(s)=s$, (\ref{eqn:minb}) yields
\begin{equation}
\Delta b^{*}=(1-\epsilon)(\mu-b)
\end{equation}
\subsubsection{ReLU}
For ReLU activation function, $f(s)=\mathrm{max}(0,s)$.  Assuming also that $\frac{\partial\mathcal{L}}{\partial x}=1$, the expectations in (\ref{eqn:minb}) correspond to the unnormalized versions of $\mathrm{0^{th}}$ and $\mathrm{1^{st}}$ moments of a truncated normal distribution with mean $\mu$ and variance $\sigma^{2}$:
\begin{equation}
\Delta b^{*} = (1-\epsilon)\left[\mu+\sigma\frac{\phi\left(\frac{-\mu}{\sigma}\right)}{1-\Phi\left(\frac{-\mu}{\sigma}\right)}-b\right]
\end{equation}
where $\phi$ and $\Phi$ are the standard normal density and distribution functions, respectively.  
\subsubsection{Sigmoid}
For sigmoid activation functions, the analysis is a bit more involved.  First, we note that in the case of sigmoid, $f(s)=(1+\mathrm{e}^{-s})^{-1}$ and $f'(s)=f(s)(1-f(s))$.  The derivative of the sigmoid function and the square of the derivative both have a bell curve shape.  Therefore, in this work, we approximate $[f'(s)]^{2}$ as a constant 
 multiplied by a normal distribution function:
\begin{equation}
[f'(s)]^{2}=\left[\frac{1}{1+e^{-s}}\left(1-\frac{1}{1+e^{-s}}\right)\right]^{2}\approx k\frac{\sqrt{2\pi}}{16}\left(\frac{1}{k}\right)\phi\left(\frac{s}{k}\right)
\end{equation}
Here, $k$ is chosen empirically to get the best approximation to  $[f'(s)]^{2}$.  We've found that $k=1.05$ gives a good result. The product of $[f'(s)]^{2}$ with $p(s)$ then becomes the product of two normal density functions, which is proportional to another normal density function
\begin{equation}
p(s)[f'(s)]^{2}\approx A\phi\left(\frac{s-B\mu}{\sqrt{B}\sigma}\right)
\end{equation}
where $A=(\sqrt{2\pi}/16)\phi\left(\mu/\sqrt{\sigma^{2}+1.05^{2}}\right)$ and $B=1.05^{2}/(\sigma^{2}+1.05^{2})$.  Plugging this expression into (\ref{eqn:minb}) gives
\begin{equation}
\Delta b^{*}\approx(1-\epsilon)(B\mu-b)
\end{equation}
\subsubsection{Hyperbolic Tangent}
For $f(s)=\mathrm{tanh}(s)$, we follow a similar procedure as in the sigmoid case, approximating $[f'(s)]^{2}$ as a normal distribution function:
\begin{equation}
[f'(s)]^{2}=\mathrm{sech}^{4}(s)\approx\sqrt{2\pi}\phi\left(\frac{s}{0.75}\right)
\end{equation}
\begin{equation}
\Delta b^{*}\approx (1-\epsilon)(B\mu-b)
\end{equation}
where $B=0.75^{2}/(\sigma^{2}+0.75^{2})$.

\bibliographystyle{ieeetr}
\bibliography{biblio.bib}
\end{document}